\def\year{2019}\relax
\documentclass[letterpaper]{article} 
\usepackage{aaai19}  
\usepackage{times}  
\usepackage{helvet}  
\usepackage{courier}  
\usepackage{url}  
\usepackage{graphicx}  

\graphicspath{{./figs/}}

\usepackage{amsmath}
\usepackage{makecell}
\usepackage{caption}
\usepackage{subcaption}
\usepackage{tabularx}
\usepackage{textcomp}

\newcommand{\citet}[1]
{\citeauthor{#1}~\shortcite{#1}}
\newcommand{\citep}{\cite}

\begin{document}
%
\title{Implicit Argument Prediction as Reading Comprehension}
\author{Pengxiang Cheng \\ Department of Computer Science \\ The University of Texas at Austin \\\texttt{pxcheng@utexas.edu}
	\And Katrin Erk \\ Department of Linguistics \\ The University of Texas at Austin \\\texttt{katrin.erk@mail.utexas.edu}
}
\maketitle

\begin{abstract}
Implicit arguments, which cannot be detected solely through syntactic cues, make it harder to extract predicate-argument tuples. We present a new model for implicit argument prediction that draws on reading comprehension, casting the predicate-argument tuple with the missing argument as a query. We also draw on pointer networks and multi-hop computation. Our model shows good performance on an argument cloze task as well as on a nominal implicit argument prediction task.
\end{abstract}

\section{Introduction}
\label{sec::intro}

Predicate-argument tuples describe ``who did what to whom'' and are an important data structure to extract from text, for example in Open Information Extraction~\citep{Etzioni:2007}. This extraction is straightforward when arguments are syntactically connected to the predicate, but much harder in the case of \emph{implicit arguments}, which are not syntactically connected to the predicate and may not even be in the same sentence. These cases are not rare; they can be found within the first few sentences on any arbitrary Wikipedia page, for example:\footnote{\url{https://en.wikipedia.org/wiki/History_of_Gillingham_F.C.}}
\begin{quote}
	Twice in the late 1980s \textit{Gillingham} came close to winning promotion to the second tier of English football, but a \underline{decline} then set in\ldots
\end{quote}
Here, \textit{Gillingham} is an implicit argument to \textit{decline}. Generally, predicates with implicit arguments can be nouns, as in the example, or verbs.

Implicit argument prediction as a machine learning task was introduced by \citet{gerber2010acl} and \citet{Ruppenhofer:2010}, and was studied in a number of papers~\citep{silberer2012sem,laparra2013acl,stern2014recognizing,chiarcos2015sigdial,schenk2016naacl,do2017ijcnlp}. In this task, the model is given a predicate-argument tuple with one or more arguments missing. The model then chooses a filler for each missing argument from the document (or chooses to leave the argument unfilled).  Building on recent work that made the task accessible to neural models through training on automatically generated data~\citep{cheng-erk-2018}, we introduce a new neural model for implicit argument prediction.

In this paper, we view the task of implicit argument prediction as related to Reading Comprehension~\citep{hermann2015teaching}: A predicate-argument tuple with the missing argument is a query. The answer to the query has to be located in the document. However the tasks are not exactly the same. 
One difference is that the answer is not a vocabulary item or text span, but a single input item. This suggests the use of Pointer Networks~\citep{vinyals2015pointer}. We obtain the Pointer Attentive Reader for implicit argument prediction. Another difference is that more than one argument may be missing in a predicate-argument tuple. In this case we want the model to reason over the whole document to derive a more informative query. We do this through a multi-hop extension, taking inspiration from multi-hop memory networks~\citep{sukhbaatar2015end}. Our model shows good performance on an argument cloze task as well as on a nominal implicit argument prediction task.

\section{Related Work}
\label{sec::related}

Recent work on implicit arguments started from \citet{gerber2010acl} and \citet{Ruppenhofer:2010}. \citet{gerber2010acl} constructed a dataset (\textbf{G\&C}) by selecting 10 nominal predicates and labeling implicit arguments of these predicates in the NomBank \citep{meyers2004nombank} corpus manually. The resulting dataset is quite small, consisting of approximately 1000 examples. They also proposed a linear classifier for the task. \citet{gerber2012cl} added more features and performed cross validation on the dataset, leading to better results.  \citet{Ruppenhofer:2010} also introduced an implicit argument dataset by annotating several chapters of fiction (\textbf{SemEval-2010}), which is even smaller (about 500 examples) and more complex than \citet{gerber2010acl}. There has since been much follow-up work proposing new methods for G\&C \citep{laparra2013acl,schenk2016naacl,do2017ijcnlp} and SemEval-2010 \citep{silberer2012sem,laparra2013iwcs,chiarcos2015sigdial}. To overcome the size limitation, several methods for creating additional training data have been proposed. \citet{pado2015sem} combined the two datasets, using one as out-of-domain training data for another. \citet{roth2015cl} identified new instances of implicit arguments by aligning monolingual comparable texts, however the size is still similar to that of G\&C and SemEval-2010. \citet{schenk2016naacl} proposed using text with automatically labeled semantic roles to learn protypical fillers. Both \citet{silberer2012sem} and \citet{cheng-erk-2018} used coreference information to obtain additional training data. \citet{silberer2012sem} used datasets with manually annotated coreference as additional training data. \citet{cheng-erk-2018} generated large amounts of training data by using automatically produced coreference labels. They also introduced an additional dataset for testing, which has manually annotated coreference \citep{Hovy2006Ontonotes} but is automatically manipulated to simulate implicit arguments. We adopt the data generation schema from \citet{cheng-erk-2018} as the scale allows training of complex neural models. We evaluate our model on the G\&C dataset, and compare to models from \citet{gerber2012cl} and \citet{cheng-erk-2018}, which have obtained the best performance on the G\&C dataset (discussed in Section \ref{sec::exp::gc}).
Recently, \citet{ogorman2018amr} introduced a new AMR corpus with annotation for more than 2000 implicit arguments. While the data is not available yet, it will significantly extend the amount of naturally occurring test data once it is available.

In this paper we draw on recent progress in reading comprehension and memory networks, 
for the task of implicit argument prediction. \citet{hermann2015teaching} first introduced neural models to reading comprehension tasks by creating a large cloze-like dataset from news articles paired with human-written summaries. They proposed an Attentive Reader model that used an attention mechanism to reason over the document and query pair. Since then there has been much follow-up work on new datasets \citep{hill2015goldilocks,rajpurkar2016emnlp,welbl2017constructing} and new models \citep{chen2016acl,seo2016bidirectional,dhingra2017acl}.
Another related line of work that is of particular interest to us is that on End-to-End Memory Networks~\citep{sukhbaatar2015end}, which use multiple layers of attention computation (called ``multiple hops'') to allow for complex reasoning over the document input.

We also draw on pointer networks in that we view implicit argument prediction as a pointer to a previous mention of an entity. \citet{vinyals2015pointer} first proposed Pointer Networks as a variant of the conventional sequence-to-sequence model that uses the attention distribution over input sequence directly as a ``pointer'' to suggest one preferred input state, instead of as a weight to combine all input states. This architecture has been applied to a number of tasks, including Question Answering \citep{xiong2017dynamic} and Machine Comprehension \citep{wang2016machine}.

\section{Task Setup}
\label{sec::task}

\begin{figure}[!ht]
	\centering
	\begin{subfigure}[b]{0.9\linewidth}
		\centering
		\includegraphics[width=\linewidth]{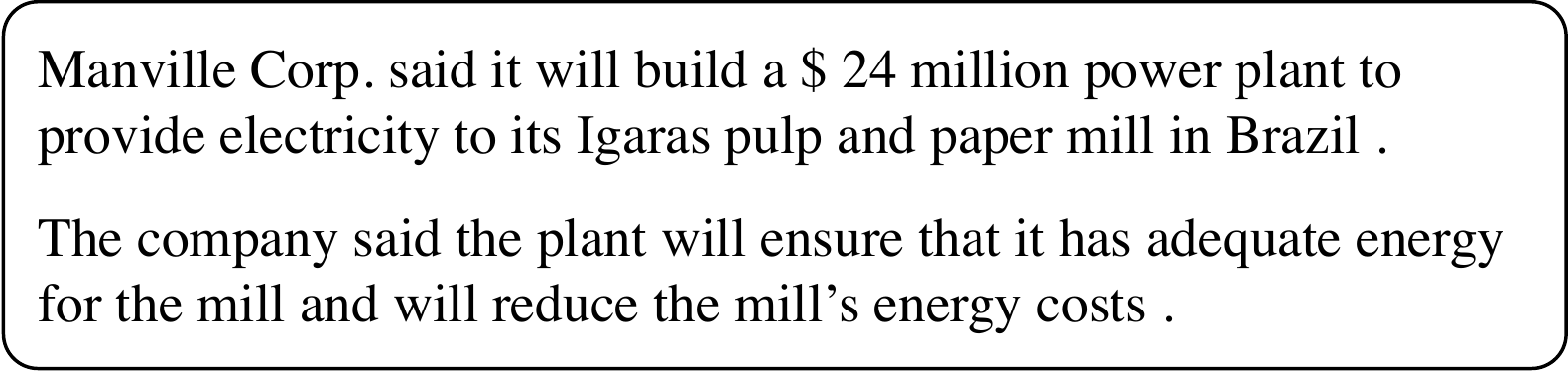}
		\caption{A piece of raw text from OntoNotes corpus.}
		\label{fig::ex-a}
	\end{subfigure}
	~
	\begin{subfigure}[b]{0.9\linewidth}
		\centering
		\includegraphics[width=\linewidth]{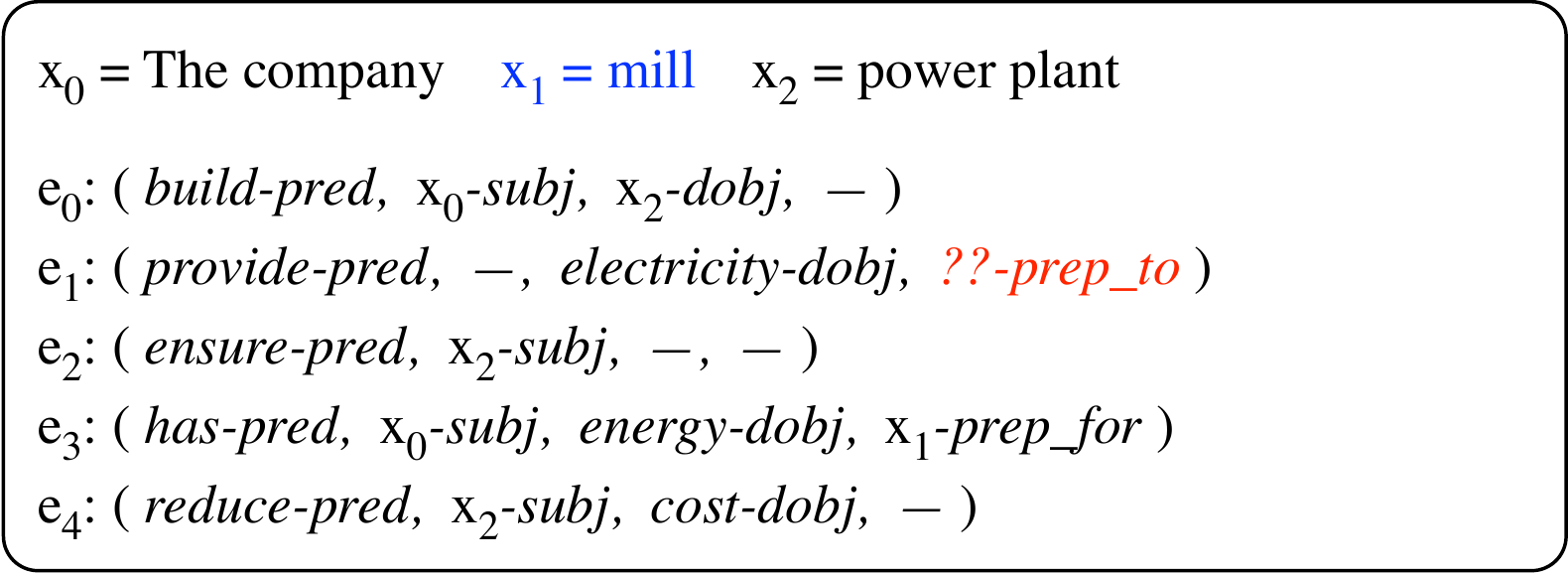}
		\caption{An example of argument cloze task.}
		\label{fig::ex-b}
	\end{subfigure}
	~
	\begin{subfigure}[b]{0.9\linewidth}
		\centering
		\includegraphics[width=\linewidth]{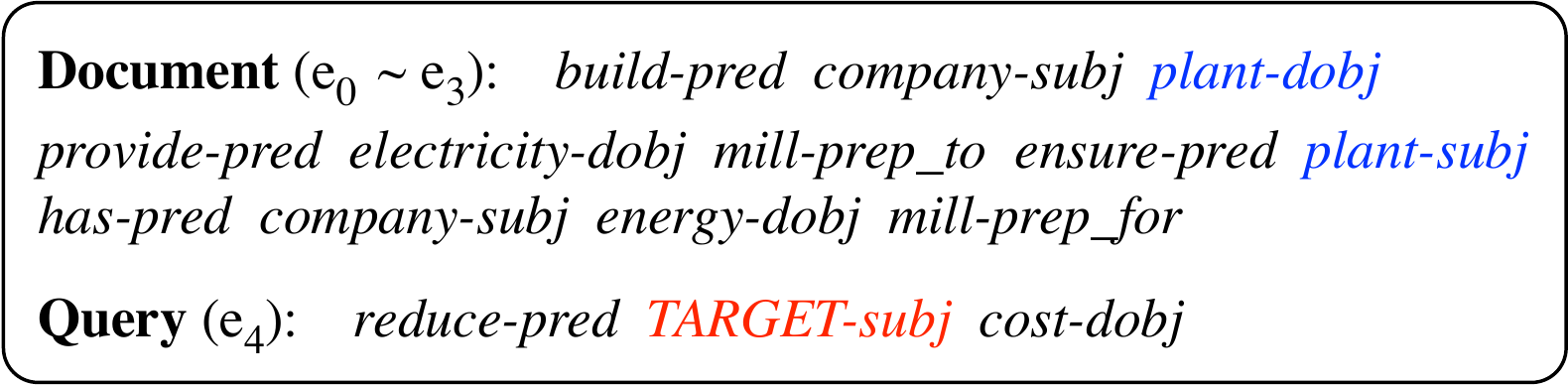}
		\caption{An example when viewed as document-query pair.}
		\label{fig::ex-c}
	\end{subfigure}
	\caption{Example of the argument cloze task and how to view it as reading comprehension. (Part \ref{fig::ex-a} and \ref{fig::ex-b} modified from \citet{cheng-erk-2018}.)}
	\label{fig::ex}
\end{figure}

The implicit argument prediction task, as first introduced by \citet{gerber2010acl}, is to identify the correct filler for an implicit argument role of a predicate, given the explicit arguments of the same predicate and a list of candidates. This task requires a lot of human effort in annotation, and the existing human-annotated datasets are too small for the use of neural models. The argument cloze task proposed by \citet{cheng-erk-2018} overcame this difficulty by automatically generating large-scale data for training. The cloze task, as shown in Figure \ref{fig::ex}, aims to simulate natural occurrences of implicit arguments, and can be briefly described as follows.

Given a piece of text with dependency parses and coreference chains ($x_0$\texttildelow $x_2$), a sequence of events (predicate-argument tuples, $e_0$\texttildelow $e_4$) are extracted from dependency relations\footnote{The event structure, in which arguments are encoded positionally after the predicate, follows common practice in recent literature of narrative schema \citep{Pichotta2016AAAI}}. Then, one argument (i.e., \emph{prep\_to} of $e_1$) that belongs to a coreference chain ($x_1$) with at least two mentions is randomly selected and removed. The model is asked to pick the removed argument from all coreference chains appearing in the text (Figure \ref{fig::ex-b}).

However, in both manually annotated  implicit argument datasets \citep{gerber2010acl,Ruppenhofer:2010}, only preceding mentions are considered as ground truth fillers, so the example in Figure \ref{fig::ex-b} is very dissimilar to naturally occurring implicit arguments. Therefore, to make the argument cloze task closer to the natural task, we change the evaluation of the task by considering candidates to be mentions, not coreference chains, and by considering only candidates that appear before the implicit argument, independent of their number of mentions.

\begin{figure*}[!htb]
	\centering
	\includegraphics[width=0.95\linewidth]{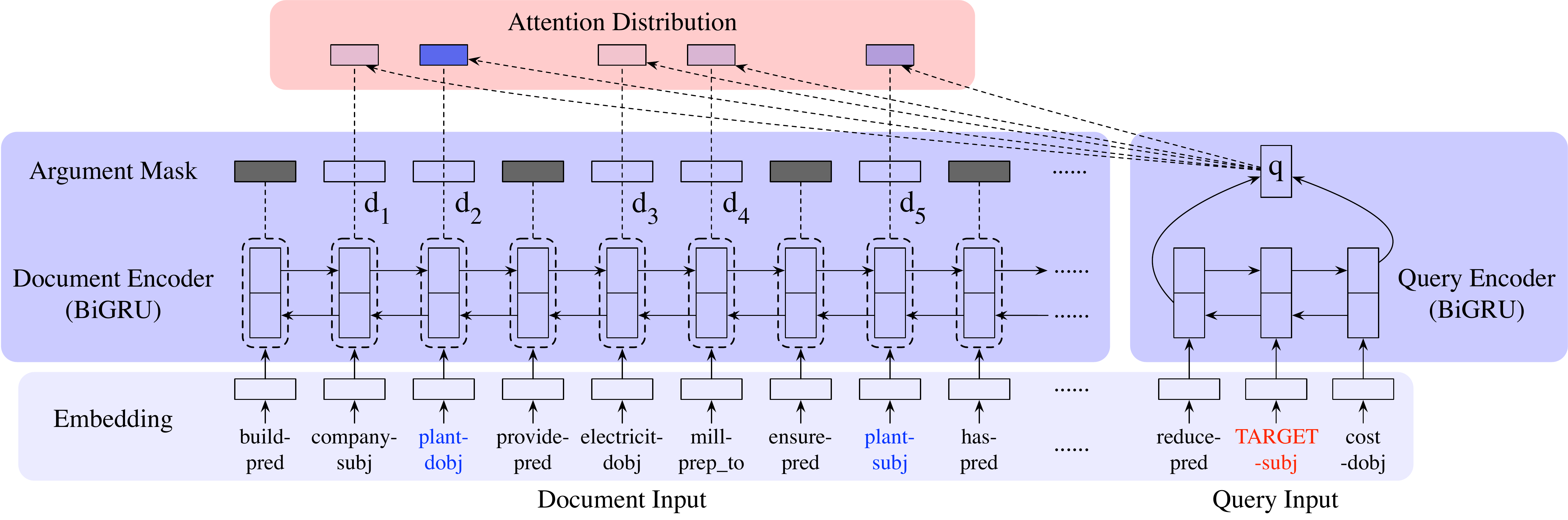}
	\caption{Pointer Attentive Reader. The \textbf{document encoder} produces a context-aware embedding for each argument mention via a BiGRU. The \textbf{query encoder}, similar to the document encoder, concatenate the last forward and backward hidden state to produce a single query vector. An \textbf{attention distribution} is computed from the query vector and all argument mention embeddings, which is then used as a pointer to select one filler for the missing argument in the query.}
	\label{fig::model}
\end{figure*}

We thus formalize the task as shown in Figure \ref{fig::ex-c}. For an event ($e_4$) with a missing argument (\emph{subj}), we concatenate the predicates and arguments of all preceding events ($e_0$\texttildelow $e_3$) and view this as the document, and we treat the target event with a special placeholder token (marked red) at the missing argument position as the query. The task is then to select any mention of the correct entity (marked blue) among the arguments appearing in the preceding document. A query may have multiple correct answers when there are multiple mentions of the removed entity, as shown in the example.

\section{Model}
\label{sec::model}

As discussed above, we view the task of implicit argument prediction as a variant of reading comprehension, in that we can treat the list of preceding events as a document and the target event with missing argument as a query. And we also draw on pointer networks and on multi-hop attention.

Most previous work on reading comprehension \citep{chen2016acl,seo2016bidirectional,dhingra2017acl} can be viewed as extending the Attentive Reader model by \citet{hermann2015teaching}. The Attentive Reader first encodes the document and the query via separate recurrent neural networks to get a list of document word vectors and one query vector. The query vector is used to obtain an attention-weighted sum over all document word vectors, which is then combined with the query vector to make the final prediction.

In the case of implicit argument prediction, however, the task is to directly select one token (an argument mention) from the document input sequence as the filler for the missing argument.
This suggests the use of Pointer Networks \citep{vinyals2015pointer}, a variant of the sequence-to-sequence model that uses the attention distribution over input states to ``point'' to a preferred input state.

So we combine the ideas from Attentive Reader and Pointer Networks and propose the \textbf{Pointer Attentive Reader (PAR)} model for implicit argument prediction, as illustrated in Figure \ref{fig::model}.

\subsection{Pointer Attentive Reader}
\label{sec::model::par}

\paragraph{Embedding}
The document input and the query input, as discussed in Section \ref{sec::task}, are both sequences of event components, represented as $[x_1^d, \dots, x_{\vert D\vert}^d]$ and $[x_1^q, \dots, x_{\vert Q\vert}^q]$ respectively (where $\vert D\vert$ and $\vert Q\vert$ are the numbers of tokens in document and query). The missing argument in the query is represented by a special placeholder token. Each token is then mapped to an embedding vector before being passed into the document encoder and query encoder. 

\paragraph{Document Encoder}
The document encoder is a bidirectional single-layer Gated Recurrent Unit (BiGRU) \citep{cho2014emnlp}. The forward and backward hidden state of each token are concatenated, with predicate tokens being masked out (as predicates are not considered as candidates), which gives us a list of context-aware embeddings of argument mentions: $[\mathbf{d_1}, \dots, \mathbf{d_T}]$.

\paragraph{Query Encoder}
The query encoder is also a BiGRU similar to the document encoder, except that we concatenate the last forward hidden state and the last backward hidden state to get the single query vector $\mathbf{q}$.

\paragraph{Attention}
For each argument mention embedding $\mathbf{d_t}$, we compute an attention score $a_t$ using the query vector $\mathbf{q}$ as\footnote{We have also tried bilinear attention and dot product attention \citep{Luong2015EMNLP}, but got lower performance.}:
\begin{align}
\label{eq::attention}
\begin{split}
s_t &=  \mathbf{v}^T\cdot \tanh(\mathbf{W} [\mathbf{d_t}, \mathbf{q}]) \\
a_t &=  \mathrm{softmax}(s_t)
\end{split}
\end{align}
where $\mathbf{W}$ and $\mathbf{v}$ are learned parameters.

Finally, the attention scores $[a_1, \dots,a_T]$ are directly used as pointer probabilities to select the most probable filler for the implicit argument.

\paragraph{Training}
Unlike conventional pointer networks where there exists a single target for the pointer, there could be multiple correct answers from the document input list in our implicit argument prediction task (as in the example in Figure \ref{fig::ex-c}). Therefore, we train the model to maximize the ``maximum correct'' attention score. That is, with a list of attention scores $\mathbf{a} = [a_1, a_2, \dots, a_T] \in\mathcal{R}^T$, and a binary answer mask $\mathbf{m_c}\in\mathcal{R}^T$ which has 1s for correct answer positions (e.g., \emph{plant-dobj} and \emph{plant-subj} in Figure \ref{fig::ex-c}) and 0s elsewhere, we train the model with the following negative log likelihood (NLL) loss function:

\begin{equation}
\label{eq::loss}
L = -\log(\max(\mathbf{a}\circ \mathbf{m_c}))
\end{equation}
where $\circ$ is element-wise multiplication.

\subsection{Multi-hop Attention}
\label{sec::model::multi-hop}

\begin{figure}[!htb]
	\centering
	\includegraphics[width=0.9\linewidth]{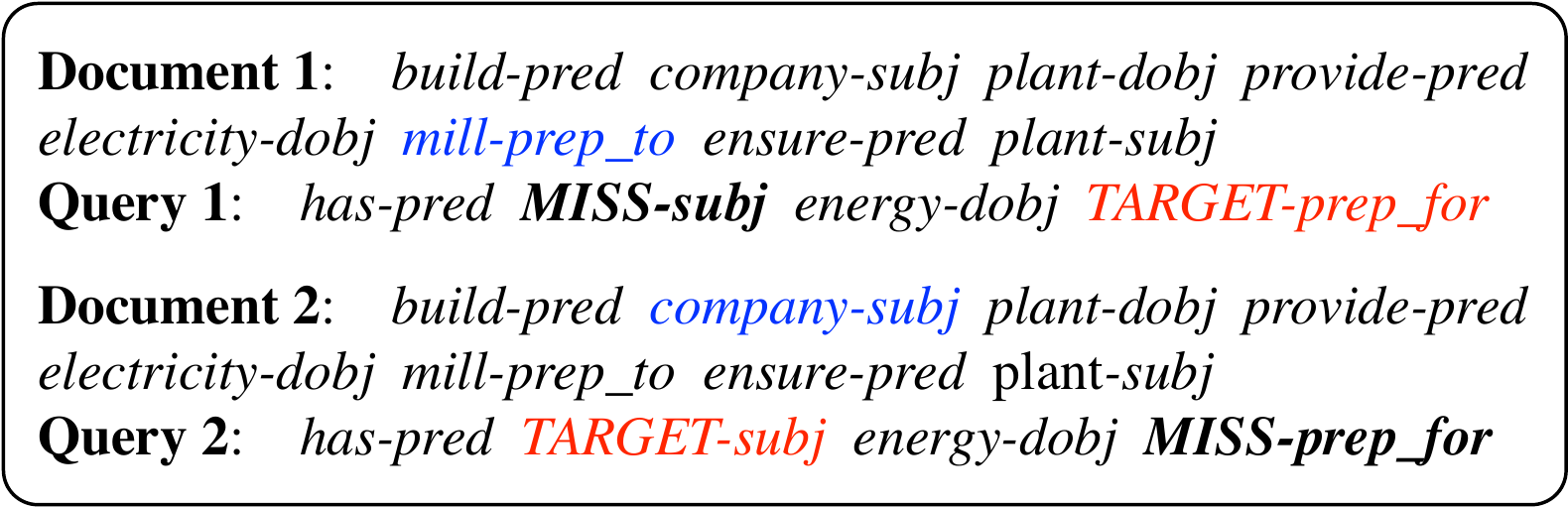}
	\caption{Document-Query example for predicates with more than one implicit argument.}
	\label{fig::ex-multi}
\end{figure}

A single event can have more than one implicit argument, and in fact this is the case for over 30\% of nominal predicates in the dataset of \citet{gerber2010acl}. In such cases, we still treat one implicit argument as the target argument to be filled, and the other arguments are indicated to the model to be missing but not target, using a separate placeholder token. An example is shown in Figure \ref{fig::ex-multi}, where target arguments are marked red, ``missing but not target'' arguments are marked bold, and answers to the target arguments are marked blue.

\begin{figure}[!htb]
	\centering
	\includegraphics[width=0.9\linewidth]{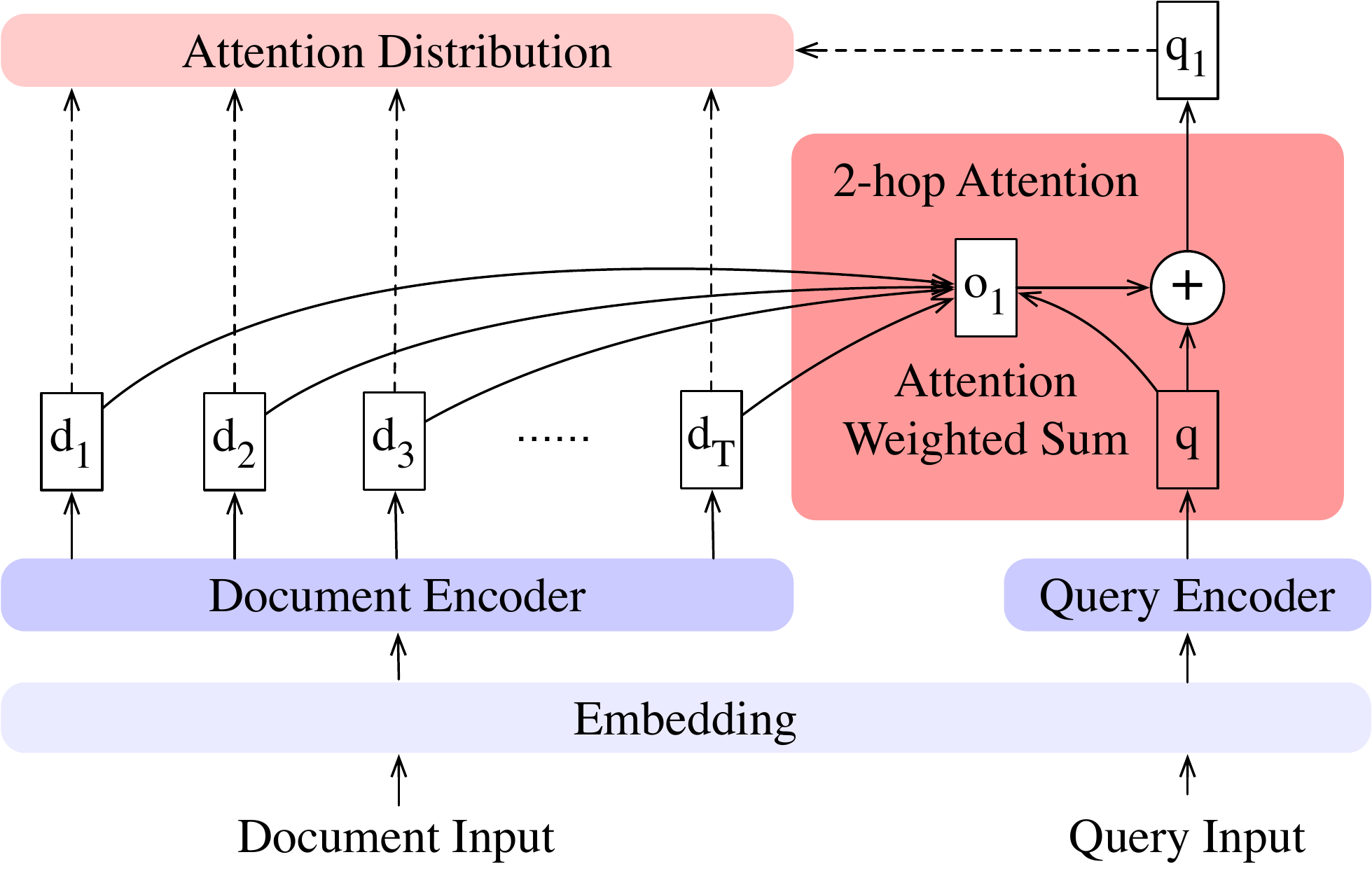}
	\caption{2-hop Pointer Attentive Reader. The query vector $\mathbf{q}$ is first updated by an attention weighted sum $\mathbf{o_1}$ from all argument embeddings in the document, before used to compute the final attention distribution.}
	\label{fig::multi-hop}
\end{figure}

When there are multiple implicit arguments, this could make the query vector $\mathbf{q}$ lack enough information to compute the correct attention distribution, especially in the extreme case where only the predicate and placeholder tokens are present in the query input. To overcome this difficulty, we strengthen the model with the ability to reason over the document and query to infer the missing but non-target arguments and thus build a better query. We do this by extending the Pointer Attentive Reader model with multi-hop attention, inspired by the idea of end-to-end memory networks \citep{sukhbaatar2015end}. For example in Figure \ref{fig::ex-multi}, we can make the vector of Query 1 more informative by attending to all missing arguments of \emph{has} in the first hop. We are not predicting the subject at this point, but could use it to help the final prediction of \emph{TARGET-prep\_for}. Figure \ref{fig::multi-hop} shows the 2-hop Pointer Attentive Reader model.

To make the query vector document-aware, we update the query vector $\mathbf{q}$, in each but the last hop, by an attention-weighted sum $\mathbf{o_1}$ over argument embeddings $[\mathbf{d_1},\mathbf{d_2},\dots,\mathbf{d_T}]$:

\begin{equation}
\label{eq::multi-hop}
\begin{aligned}[c]
s_t^\prime &=  {\mathbf{v^\prime}}^T\cdot \tanh(\mathbf{W^\prime}[\mathbf{d_t}, \mathbf{q}]) \\
a_t^\prime &=  \mathrm{softmax}(s_t^\prime) \\
\mathbf{o_1} &= \Sigma_{t=1}^{T}a_t^\prime\cdot \mathbf{d_t} \\
\mathbf{q_1} &= \mathbf{o_1} + \mathbf{q}
\end{aligned}
\end{equation}
where $\mathbf{W^\prime}$ and $\mathbf{v^\prime}$ are learned parameters. Then in Equation \ref{eq::attention} we use $\mathbf{q_1}$ instead of $\mathbf{q}$ to compute the final attention scores.

In this paper we only experiment with 2-hop attention. However the model can be easily extended to $k$-hop ($k > 2$) attention models.

\paragraph{Extra Supervision} Another advantage of using multi-hop attention is that we can apply extra supervision \citep{hill2015goldilocks} on the attention scores to force the model to learn any arbitrary attention distribution as desired. In the case of multiple implicit arguments, we want the model to attend to all missing arguments of the query event in the first hop of attention, so that the query vector receives enough information for subsequent hops. Therefore, the desired distribution has $1/k$ for all mentions of all missing arguments (assuming $k$ mentions in total) and 0 elsewhere. (In the examples in Figure \ref{fig::ex-multi}, the distribution would have $0.5$ for both \emph{company-subj} and \emph{mill-prep\_to}.) Then we can add the KL-divergence between the actual attention scores and the desired distribution to the loss function in Equation \ref{eq::loss}.

\section{Empirical Results}
\label{sec::exp}

\subsection{Training Data}
\label{sec::exp::implementation}

\paragraph{Preprocessing} We construct a large scale training dataset from the full English Wikipedia corpus. We retrieve each document from the 20160901 dump of English Wikipedia\footnote{\url{https://dumps.wikimedia.org/enwiki/}}, and split it into paragraphs using the WikiExtractor tool\footnote{\url{https://github.com/attardi/wikiextractor}.}.

We run Stanford CoreNLP \citep{Manning2014ACL} to obtain dependency parses and coreference chains of each paragraph,\footnote{The coreference chains are used to make training data, but are not handed to the model.} from which we extract a sequence of events and entities as demonstrated in Figure \ref{fig::ex-b}, after lemmatizing all verbs and arguments, incorporating negation and particles to verbs, and normalizing passive constructions. We downsample the most frequent verbs (with counts over 100,000) by a ratio proportional to the square root of their counts, then construct a document-query pair for every argument of every event in the sequence if the argument co-refers with at least one argument in its preceding events (Figure \ref{fig::ex-c}). This leads to approximately 25 million document-query pairs in the training dataset. This dataset is used to train models for both evaluation tasks discussed below.

\paragraph{Initialization and Hyperparameters} For training the Pointer Attentive Reader model, we initialize the embedding layer with event-based word2vec embeddings, following \citet{cheng-erk-2018}. (The embedding vectors for placeholder tokens are initialized to zero.) We use a hidden size of 300 in both document encoder and query encoder, and apply a dropout layer with a rate of 0.2 on all embeddings before they are passed to the encoders. We train the model for 10 epochs with a batch size of 128, using Adagrad optimizer \citep{duchi2011adaptive} to minimize the negative log-likelihood loss as defined in Equation \ref{eq::loss} with a learning rate of 0.01. The 2-hop Pointer Attentive Reader model is trained with the same set of hyperparameters.

\subsection{Evaluation on OntoNotes Dataset}
\label{sec::exp::synthetic}

Our main evaluation is on the argument cloze task using the OntoNotes datasets of \citet{cheng-erk-2018}. The datasets are large and provide clean test data, as they are based on gold syntax and coreference annotation.  The two datasets are \textsc{On-Short} and \textsc{On-Long}, where the latter consists of considerably longer documents. We modify their data generation pipeline\footnote{\url{https://github.com/pxch/event_imp_arg}}  as discussed in Section \ref{sec::task}. This greatly reduces the number of test cases, as many cases in the original setting have the missing argument only coreferring with arguments of subsequent events, which are excluded in our new setting. Also, although now there can be more than one candidate that constitutes a correct answer to a query (as in the example in Figure \ref{fig::ex-c}), the number of candidates also grows much larger (about three times), because we now view every argument mention rather than a whole coreference chain as a candidate. Some statistics of both the original and modified datasets are shown in Table \ref{tab::synthetic-data}.

\begin{table}[!htb]
	\small
	\centering
	\begin{tabularx}{\linewidth}{c c c c c}
		\hline
		& \multicolumn{2}{c}{\textsc{ON-Short}} & \multicolumn{2}{c}{\textsc{ON-Long}} \\
		\cline{2-5}
		& Original & Modified & Original & Modified \\
		\hline
		\# doc & \multicolumn{2}{c}{1027} & \multicolumn{2}{c}{597} \\
		\# test cases & 13018 & 7781 & 18208 & 10539 \\
		Avg \# candidates & 12.06 & 34.99 & 36.95 & 93.89 \\
		Avg \# correct & 1 &  3.17 & 1 & 4.61 \\
		\hline
	\end{tabularx}
	\caption{Statistics of the  OntoNotes datasets.}
	\label{tab::synthetic-data}
\end{table}

We compare our model to 2 baselines, the \textbf{\textsc{Random}} baseline, which randomly selects one candidate, and the \textbf{\textsc{MostFreq}} baseline, which selects any candidate belonging to the coreference chain with highest number of mentions. We also compare with the best performing \textbf{\textsc{EventComp}} model in \citet{cheng-erk-2018}.

\paragraph{Results} The evaluation results are shown in Table \ref{tab::eval-ontonotes}. We can see that the Pointer Attentive Reader outperforms the previously best \textsc{EventComp} model by a large margin, especially on the harder \textsc{On-Long} dataset. \citet{cheng-erk-2018} found that entity salience features, that is, numbers of different types of mentions in a coreference chain, greatly improves the performance of their \textsc{EventComp} model. We have also tried to add such features to our model, but do not see significant improvement (sometimes adding the features even degrades the performance). This is probably due to the fact that by sequentially modeling the context through a document encoder, PAR is already encoding entity salience as some latent information in its context-aware vectors $[\mathbf{d_1}, \dots, \mathbf{d_T}]$.

\begin{table}[!htb]
	\small
	\centering
	\begin{tabular}{l c c}
		\hline
		& \textsc{On-Short} & \textsc{On-Long} \\
		\hline
		\textsc{Random} & 13.24 & 8.74 \\
		\textsc{MostFreq} & 35.15 & 26.29 \\
		\hline
		\textsc{EventComp} & 36.90 & 21.26 \\
		\quad + entity salience & 46.06 & 31.43 \\
		\hline
		\textsc{PAR} & \textbf{58.12} & \textbf{51.52} \\
		\hline
	\end{tabular}
	\caption{Evaluation on the OntoNotes datasets.}
	\label{tab::eval-ontonotes}
\end{table}

\begin{figure}[!htb]
	\centering
	\includegraphics[width=0.9\linewidth]{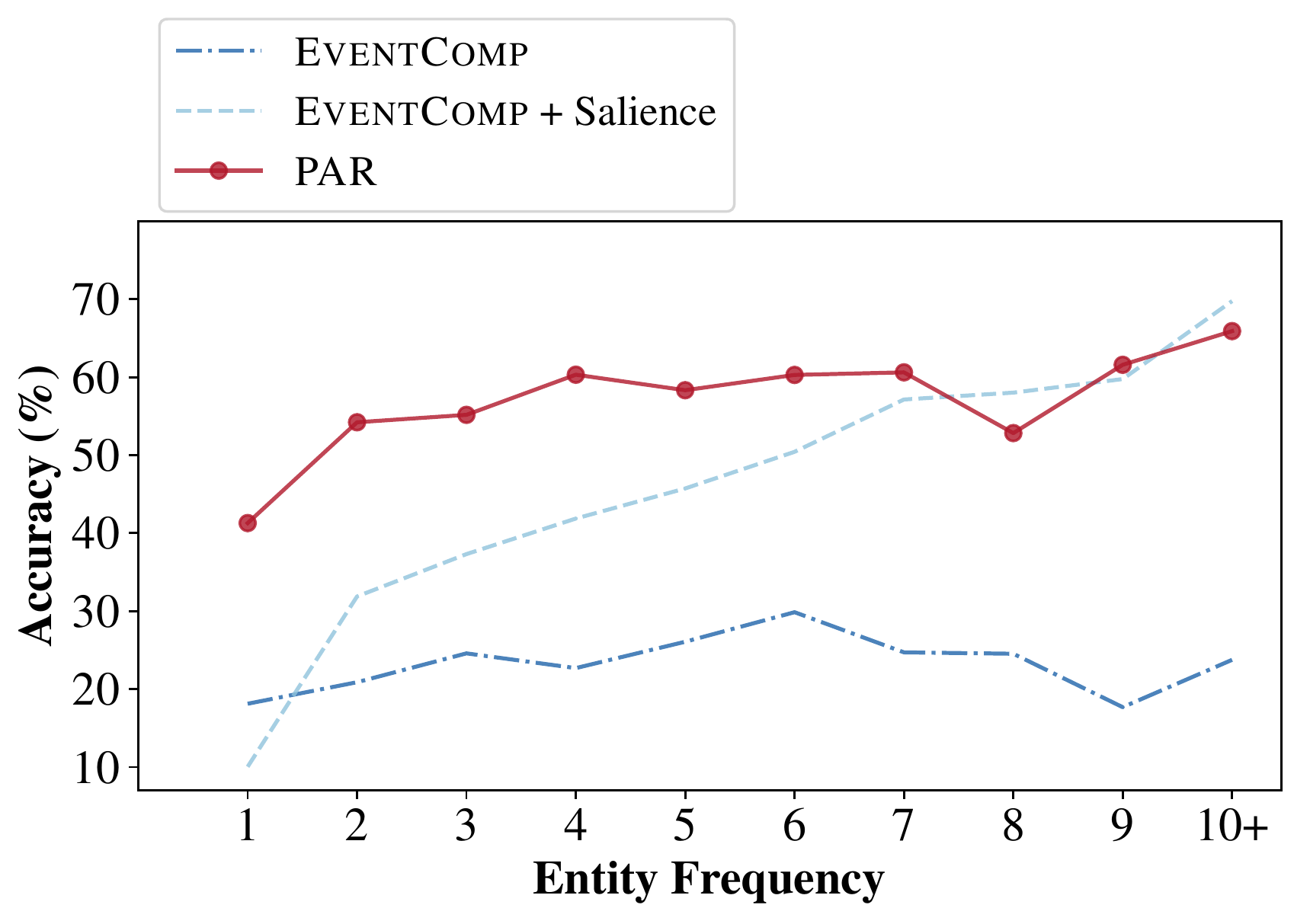}
	\caption{Performance of \textsc{EventComp}, with and without entity salience, and \textsc{PAR}, by entity frequency (length of coreference chain) of the removed argument, on \textsc{On-Long}.}
	\label{fig::case-analysis}
\end{figure}

To better understand why PAR is performing well, we plot the accuracy of different models on \textsc{On-Long} by the frequency of the removed argument, that is, by the number of preceding mentions referring to the argument, in Figure \ref{fig::case-analysis}. We can see that entity salience boosts the performance of the \textsc{EventComp} model in particular for frequent entities. While PAR not only achieves comparable performance on frequent entities with \textsc{EventComp} + entity salience, it also maintains a relatively steady performance on rare entities, indicating that our model is able to capture both semantic content of events and salience information of entities.

\begin{figure*}[!htb]
	\centering
	\includegraphics[width=0.95\linewidth]{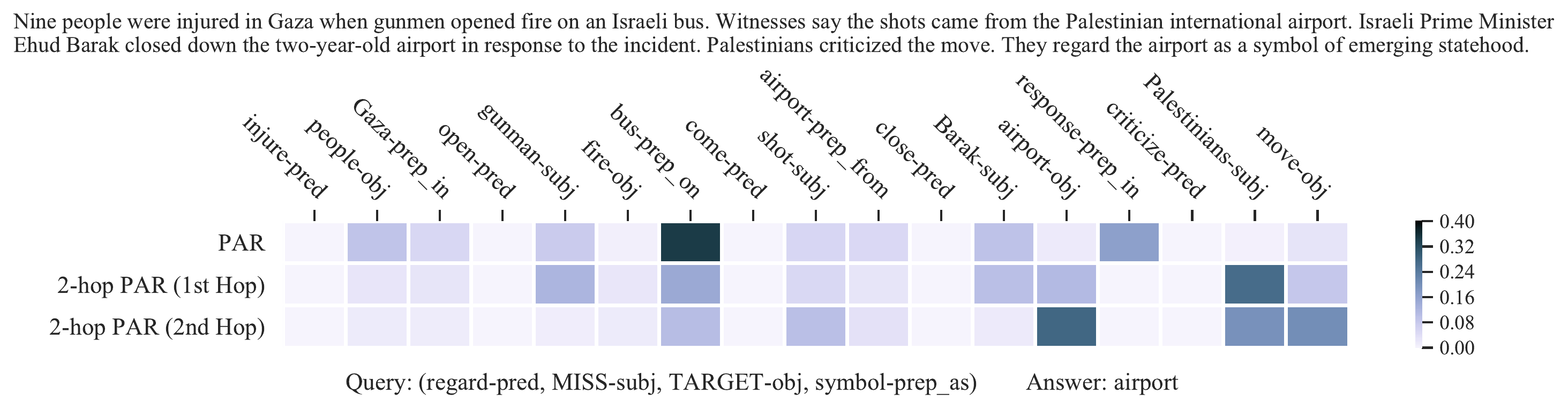}
	\caption{An example from the OntoNotes dataset (\texttt{english/bn/cnn\_0019}) with multiple implicit arguments, and the attention scores computed by PAR and 2-hop PAR. While PAR fails on this example, 2-hop model succeeds from a more informative query vector when the first hop attends to other missing arguments of the query.}
	\label{fig::heatmap}
\end{figure*}

\paragraph{Evaluation on Multiple Implicit Arguments}

To test our model's ability to predict multiple implicit arguments of the same predicate (Section \ref{sec::model::multi-hop}), we extract subsets from both the \textsc{On-Short} and \textsc{On-Long} datasets by selecting queries with more than one argument that is a potential implicit argument (i.e., co-referring with arguments of preceding events). Then we modify each query by removing all such potential implicit arguments, and ask the model to predict one of them at a time, as in the examples shown in Figure \ref{fig::ex-multi}. We name the resulting two subsets \textsc{On-Short-Multi} and \textsc{On-Long-Multi}.

\begin{table}[!htb]
	\small
	\centering
	\begin{tabular}{l c c}
		\hline
		& \makecell[cl]{\textsc{On-Short}\\\textsc{-Multi}} & \makecell[cl]{\textsc{On-Long}\\\textsc{-Multi}} \\
		\hline
		\textsc{PAR} w/o multi-arg & 51.49 & 43.06 \\
		\hline
		\textsc{PAR} & 48.45 & 39.90 \\
		\textsc{2-hop PAR} & 50.54 & \textbf{42.69} \\
		\quad + extra supervision & \textbf{50.73} & 41.72 \\
		\hline
	\end{tabular}
	\caption{Evaluation on subsets of the OntoNotes datasets with more than one missing argument in the query.}
	\label{tab::eval-ontonotes-multi}
\end{table}

Table \ref{tab::eval-ontonotes-multi} shows the result of testing PAR and 2-hop PAR on the two subsets. The ``\textsc{PAR} w/o multi-arg'' evaluates PAR on the same subsets of queries, but only removes one argument at a time. The performance drop of over 3 points from the same model proves that the multi-argument cases are indeed harder than single-argument cases. The 2-hop model, however, brings the performance on multi-argument cases close to single-argument cases. This confirms our hypothesis that multi-hop attention allows the model to build a better query by reasoning over the document.
We also trained a 2-hop model with extra supervision on the first hop of attention scores, as discussed in Section \ref{sec::model::multi-hop}, but it does not provide much benefit in this experiment. 
Figure \ref{fig::heatmap} shows an example where PAR fails to point to the correct answer, but the 2-hop model succeeds by first attending to other missing arguments of the query (\emph{Palestinians} as missing subject) in the first hop, then pointing to the correct answer in the second hop with a more informative query vector.

\subsection{Evaluation on G\&C Dataset}
\label{sec::exp::gc}

The implicit argument dataset by Gerber and Chai (2010 2012) is a very small dataset with 966 annotated implicit arguments, comprising only 10 nominal predicates. Still it is currently the largest available dataset of naturally occurring implicit arguments.

The task is, for each missing argument, to either choose a filler from a list of candidates or to leave the argument unfilled. The candidates for each missing argument position consists of all core arguments labeled by PropBank \citep{palmer2005propbank} or NomBank \citep{meyers2004nombank} within a two-sentence candidate window (i.e., the current sentence and the preceding two sentences). An example is shown below:

\begin{quote}
	The average interest rate rose to 8.3875\% at [Citicorp]\textsubscript{\emph{subj}} 's \$50 million weekly auction of [91-day commercial paper]\textsubscript{\emph{obj}}, or corporate IOUs, from 8.337\% at last week's [sale]\textsubscript{\emph{pred}}.
\end{quote}
where \emph{Citicorp} is the implicit subject of \emph{sale}, and \emph{91-day commercial paper} is the implicit object of \emph{sale}.

There are two obstacles to applying our Pointer Attentive Reader to the task.
First, the number of missing argument positions (3737) is much larger than the number of gold implicit arguments, making the dataset highly biased. Whether a particular argument position is typically filled is mostly predicate-specific, and the size of dataset makes it hard to train a complex neural model. This problem was also noted by \citet{cheng-erk-2018}, who trained a simple fill / no-fill classifier with a small subset of shallow lexical features used originally by \citet{gerber2012cl}. We adapt the same idea to overcome the problem. This also makes our results comparable to \citet{cheng-erk-2018}.

Second, our model only considers arguments of preceding verbal events (i.e., with verb predicates) as candidates. However, many of the candidates defined by the task, especially those from NomBank annotations, are not present in any verbal event (arguments of nominal predicates are likely to be absent from any dependency relation with a verb). To make a fair comparison, we convert every NomBank proposition within the candidate window to an event by mapping the nominal predicate to its verbal form, and add it to the list of preceding events. After adding the extra events, there still remains a slight difference between the candidates available to our PAR model and the candidates defined by the task, which we adjust by masking out the unavailable candidates from other models used in comparison.

\paragraph{Cross Validation} The Wikipedia training data for our Pointer Attentive Reader contains only verbal predicates, and the text is from a different domain than the G\&C dataset. To bridge the gap, we fine tune the model on G\&C dataset by 10-fold cross validation, that is, for each testing fold, the model is tuned on the other nine folds. We remove the dropout layers in both document encoder and query encoder to ensure reproducibility. To prevent overfitting, we freeze the parameter weight in embedding layer and query encoder layer, using Adagrad optimizer with a learning rate of 0.0005. Still, due to the size of the dataset and the complexity of the model, the performance is very sentitive to other hyperparameters, and we cannot find a single set of hyperparameters that works best for all models. Therefore, we report our results as an average of 5 runs with slightly different hyperparameter settings. \footnote{The hyperparameters are: $(B=4, \lambda=1.0)$, $(B=8, \lambda=1.0)$, $(B=16, \lambda=1.0)$, $(B=8, \lambda=0.1)$, and $(B=8, \lambda=0.0)$, where $B$ is the batch size and $\lambda$ is the $\ell_2$ regularizer weight.}

\paragraph{Results} The evaluation results are presented in Table \ref{tab::eval-gerber-chai}. The \textsc{GCAuto} and \textsc{EventComp} results are from \citet{cheng-erk-2018};  \textsc{GCAuto} is a reimplementation of \citet{gerber2012cl} without gold features. \textsc{EventComp*} evaluates the \textsc{EventComp} model in a condition that masks out some candidates to make it a fair comparison with our PAR model, as discussed above. Note that \textsc{GCAuto}, \textsc{EventComp} and \textsc{EventComp*} all have an intrinsic advantage over the PAR model as they exploit event information from the whole document to make the prediction, while our new model only looks at the preceding text.

\begin{table}[!htb]
	\small
	\centering
	\begin{tabular}{l l l l}
		\hline
		& $P$ & $R$ & $F_1$ \\
		\hline
		\citet{gerber2012cl} & 57.9 & 44.5 & 50.3 \\
		\textsc{GCauto} & 49.9 & 40.1 & 44.5 \\
		\textsc{EventComp} & 49.3 & 49.9 & 49.6 \\
		\hline
		\textsc{EventComp*}  & 48.0 & 48.7 & \textbf{48.3} \\
		\hline
		\textsc{PAR} & 44.0 & 44.7 & 44.4 \\
		\textsc{2-Hop PAR} & 45.9 & 46.6 & 46.2 \\
		\quad + extra supervision & 47.9 & 48.6 & \textbf{48.3} \\
		\hline
	\end{tabular}
	\caption{Evaluation on the \textsc{G\&C} dataset.}
	\label{tab::eval-gerber-chai}
\end{table}

\begin{figure}[!htb]
	\centering
	\includegraphics[width=\linewidth]{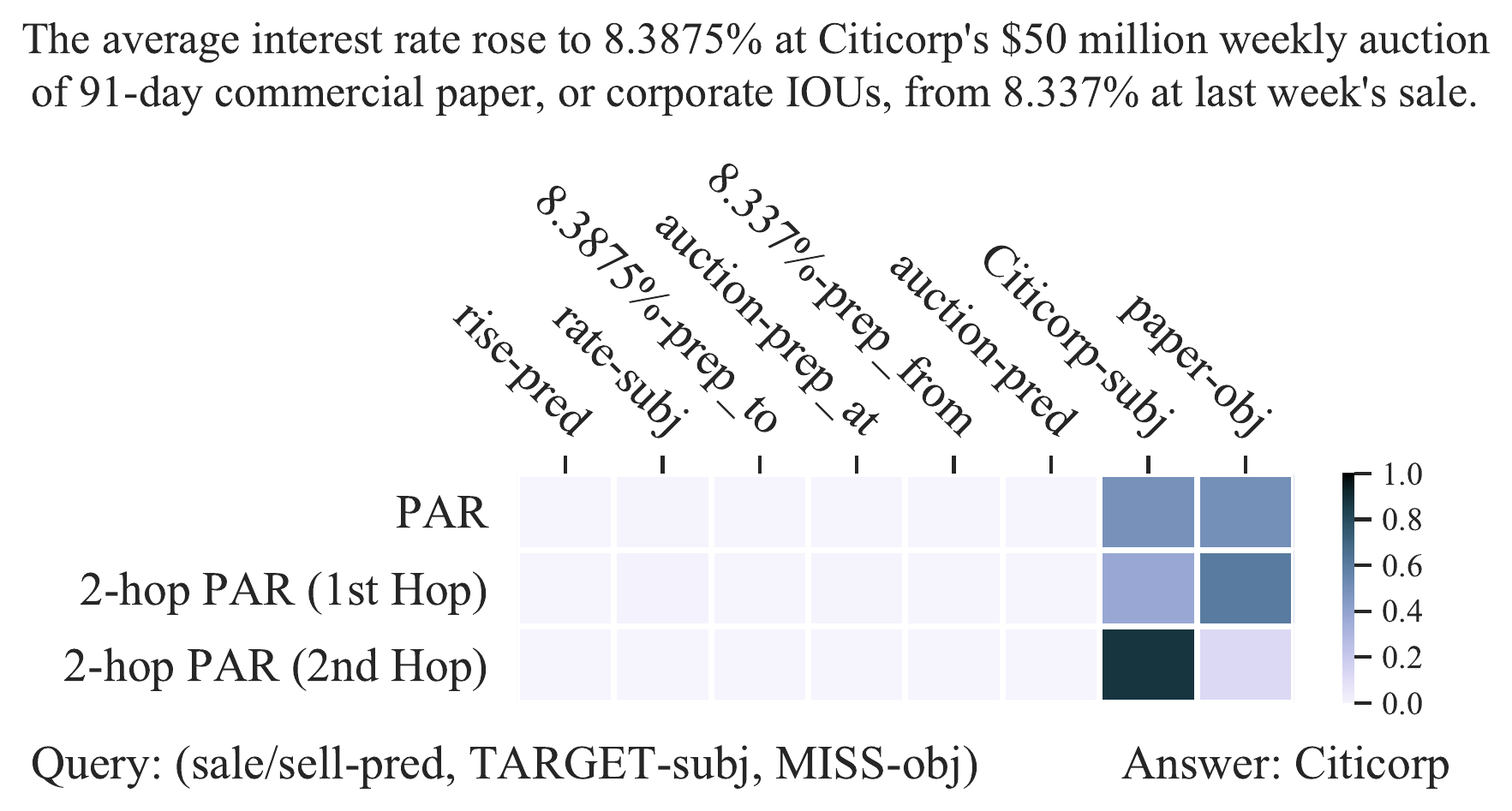}
	\caption{A G\&C example with multiple implicit arguments, and the attention scores computed by PAR and 2-hop PAR. While the 2-hop model attends more to the non-target missing argument (paper-obj) on the first hop, it successfully points to the target argument in the second hop.}
	\label{fig::heatmap-2}
\end{figure}

The performance of the plain \textsc{PAR} model is already comparable to the \textsc{GCAuto} baseline. With an additional hop of attention, the performance increases by around 2 points. This is as expected, as over 30\% of the predicates in the G\&C dataset have more than one implicit argument, and we have shown in Section \ref{sec::exp::synthetic} that multi-hop attention helps prediction on multi-argument cases. Finally, when the 2-hop model is trained with extra supervision, it gains another 1.7 points improvement, achieving an F1 score of 48.3, on par with \textsc{EventComp*}, the comparably evaluated \textsc{EventComp}. Figure \ref{fig::heatmap-2} shows the attention scores of PAR and 2-hop PAR on the previous example, to demonstrate the power of 2-hop inference on multi-argument cases.

\section{Conclusion}
\label{sec:conclusion}

In this paper we have framed implicit argument prediction as a reading comprehension task, where the predicate-argument
tuple with the missing argument is a query, and the preceding text is
the document in which the answer can be found. Also drawing on pointer
networks and multi-hop memory networks, we have introduced the Pointer
Attentive Reader model for implicit argument prediction. On an
argument cloze task, the Pointer Attentive Reader beats the previous
best model by a large margin, showing good performance on short and
long texts, and on salient as
well as less salient arguments. When multiple arguments are
missing, the use of a second hop
to reason over possible arguments of the query considerably improves
performance. This also proves useful on a small dataset of naturally 
occurring nominal predicates. Our code is available at \url{https://github.com/pxch/imp_arg_rc}.

In this paper, we have formulated the implicit argument prediction as a
task of selecting a mention of an argument, ignoring coreference. In
future work, we plan to adapt other widely used reading comprehension models, 
like BiDAF \citep{seo2016bidirectional}, to our task. Another interesting direction is to 
model coreference latently, through self-attention during the computation of embeddings
for the document. We are also interested in integrating implicit
argument reasoning in actual reading comprehension. Because 
argument cloze can be viewed as a variant of reading comprehension, models
trained on argument cloze can be straightforwardly integrated into
models for reading comprehension.

\section*{Acknowledgments}
This research was supported by NSF grant IIS 1523637 and by the DARPA AIDA program under AFRL grant FA8750-18-2-0017. We acknowledge the Texas Advanced Computing Center for providing grid resources that contributed to these results, and some results presented in this paper were obtained using the Chameleon testbed supported by the National Science Foundation. We would like to thank the anonymous reviewers for their valuable feedback.

\fontsize{9.0pt}{10.0pt} \selectfont
\bibliographystyle{aaai}
\bibliography{ref}

\end{document}